# Hardware Implementation of Spiking Neural Networks Using Time-To-First-Spike Encoding

Seongbin Oh, Dongseok Kwon, Gyuho Yeom, Won-Mook Kang, Soochang Lee, Sung Yun Woo, Jang Saeng Kim, Min Kyu Park and Jong-Ho Lee*

Department of Electrical and Computer Engineering (ECE) and Inter-University Semiconductor Research Center (ISRC), Seoul National University, Seoul, Korea

* Correspondence:
Jong-Ho Lee
jhl@snu.ac.kr

## Abstract

Hardware-based spiking neural networks (SNNs) are regarded as promising candidates for the cognitive computing system due to low power consumption and highly parallel operation. In this work, we train the SNN in which the firing time carries information using temporal backpropagation. The temporally encoded SNN with 512 hidden neurons showed an accuracy of 96.90% for the MNIST test set. Furthermore, the effect of the device variation on the accuracy in temporally encoded SNN is investigated and compared with that of the rate-encoded network. In a hardware configuration of our SNN, NOR-type analog memory having an asymmetric floating gate is used as a synaptic device. In addition, we propose a neuron circuit including a refractory period generator for temporally encoded SNN. The performance of the 2-layer neural network consisting of synapses and proposed neurons is evaluated through circuit simulation using SPICE. The network with 128 hidden neurons showed an accuracy of 94.9%, a 0.1% reduction compared to that of the system simulation of the MNIST dataset. Finally, the latency and power consumption of each block constituting the temporal network is analyzed and compared with those of the rate-encoded network depending on the total time step. Assuming that the total time step number of the network is 256, the temporal network consumes 15.12 times lower power than the rate-encoded network and can make decisions 5.68 times faster.

## 1    Introduction

Recently, artificial neural networks (ANNs) have shown remarkable results surpassing humans in some tasks such as pattern recognition, object detection, or natural language processing [1-5]. The success of ANN has attributed to the multi-layered structure inspired by the nervous system and computing the complex nonlinear transformations [6,7]. Conventional ANN, however, has fundamentally different structures from the human brain in that the time has no effect on the propagation of data and uses analog-valued neurons [8]. Also, software-based ANNs are far from real-time and low power processing, making computing on the edge devices is challenging. In this perspective, SNNs using analog synaptic devices are regarded as a greatly competent network. In SNN, data propagates in the form of short spikes as in biological neural system [9,10]. Such short pulses perform a read operation on each synaptic device, and the total current flowing in the array is integrated into the analog neuron by Kirchhoff's rule, allowing high-performance parallel computation such as Vector-by-Matrix multiplication (VMM).

There are several methods to encode the input data of multiple resolutions into the input pulse train of SNN. Commonly, the rate of pulses can be proportional to the intensity of the input data. In the rate-encoded network, the I&F behavior of the neuron is almost matched to the ReLU activation function [11]. Therefore, the weights trained by ANNs can be used directly in SNNs, and these networks have shown great performance on the complex benchmark such as CIFAR [12] or ImageNet [11,13]. However, encoding an analog input value in the form of a firing rate requires a large number of spikes to express the intensity of one input data. The rate-encoding method needs to be improved for efficient computing on the edge devices in terms of power consumption and latency.

Another candidate for encoding method is temporal encoding, where the input data is transformed to the firing time of the input spikes [14]. Temporally encoded networks can generate a single spike at most regardless of the intensity of the data so that the firing of the neuron can be sparse. There have been several efforts to train the networks encoded by the firing time of spikes. However, many works used complex synaptic functions, which create an additional burden when implemented in hardware [15-18]. Also, the system of some other works is not power-efficient due to its long duration of input pulses, not a spike [19].

In this paper, we configure SNN at the circuit level, where information is carried as the firing time of a single spike by adopting the temporal encoding method. First, by using temporal backpropagation algorithm [20], we evaluate the performance of SNN at a system-level on MNIST data sets and investigate the non-ideal issues that can occur in a hardware implementation. Afterward, we propose neuron circuit blocks to generate a refractory period for a single spike-SNN. By combining proposed neuron circuits with the synaptic device reported from our previous work, the full network is simulated at a circuit level using HSPICE. Finally, the power consumed by each block and the latency of the network are analyzed and compared with that of a rate-encoded network with the same size.

## 2  Methods

### 2.1  Training Algorithm

### 2.1.1 Forward Propagation

In this network, the input information of the SNN is encoded using the time-to-first-spike (TTFS) method. As shown in eq. (1), the firing time of input neurons are inversely proportional to the input value ($I_i$) of each individual neuron [20-24]. The more salient features spike earlier, and the less input value corresponds to a late spike. Input neurons corresponding to data 0 fires at the last time step.

$$t_i^{input} = \left\lfloor \frac{I_{max} - I_i}{I_{max}} T_{max} \right\rfloor. \tag{1}$$

Fig. 1 (a) depicts a schematic diagram of SNN encoded by TTFS method. The firing time of the $j$th neuron in the $l$th layer is defined as $t_j^l$. Input neurons emit a single spike only at their own firing time. $x_j^l(t)$ represents the input spikes generated by the $j$th neuron of $l$th layer at time $t$ in the form of voltage pulse. Input pulses are multiplied by weights and integrated by the non-leaky IF model. As expressed in eq. (3), when the membrane voltage ($V_{mem,j}^l(t)$) reaches the neuron threshold ($V_{th}^l$), the neuron fires and generates a spike ($x_j^l(t) = 1$) in the next layer. Then, the firing time of neuron, $t_j^l$, is set to $t$, the time when the membrane reaches the threshold. We assume that each neuron can generate at most a single spike per each image because the fired neuron enters the refractory period and no longer responds to the image.

$$if\ l = 1\ \ :\ x_j^l(t) = \begin{cases} 1 & (if\ t = t_j^l) \\ 0 & (o.w.) \end{cases} \tag{2}$$

$$else\ (l \neq 1):\ x_j^l(t) = \begin{cases} 1 & (if\ V_{mem,j}^l(t) > V_{th}^l\ \&\ S_j^l(t-1) = 0) \\ 0 & (o.w.) \end{cases} \quad (3)$$

where $S_j^l(t)$ denotes the cumulative input function of $j$th neuron at time $t$, which is a parameter indicating whether the neuron is a fired state at time $t$. As shown in eq. (5), the membrane voltage of output neuron $j$ can be calculated by multiplying cumulative input and weights:

$$S_j^l(t) = \begin{cases} 1 & (if\ t \geq t_j^l) \\ 0 & (if\ t < t_j^l) \end{cases} \quad (4)$$

$$V_{mem,k}^{l+1}(t) = V_{mem,k}^{l+1}(t-1) + \sum_j^{N^l} x_j^l(t) w_{jk}^l = \sum_j^{N^l} S_j^l(t) w_{jk}^l \quad (5)$$

### 2.1.2 Backward Propagation

First, note that the backward propagation process is a modified form of the algorithm of the previous work [20]. In TTFS network, the output value of neuron $k$ is expressed as the firing time ($t_k^o$). Accordingly, we defined the error function as defined in eq. (6) and (7), so that the output neuron can fire as close as possible to the target firing time of each neuron ($T_{target,k}$).:

$$L = \frac{1}{2} \sum_k^{N^o} e_k^2 \quad (6)$$

$$if\ l = output:\ \delta_k^o = e_k = (T_{target,k} - t_k^o)/T_{max}. \quad (7)$$

In order to compute the change of the weight through backpropagation, the derivative of the firing time respect to its membrane potential is required. It is not clear to calculate this value corresponding to the activation function of a typical ANN. Instead, we conducted Monte Carlo simulation using various images of MNIST data sets to model the relationship as shown in Figure 1 (b). The $x$-axis of Fig. 1 (b) represents the value of the weighted sum, which is the product of input and weights. The $y$-axis denotes the firing time ($t_j^l$) of the neuron, which is the time when the value of $x$-axis exceeds $V_{th}^l$. The result of simulation in multiple MNIST data sets has shown that the derivative can be approximated to -1 if the firing time is not 0 or $T_{max}$. On the other hand, if the $V_{mem}$ does not exceed the neuron threshold ($t_j^l = T_{max}$) or reach too quickly ($t_j^l = 0$), the firing time does not change with $V_{mem}$, so the derivatives are assumed to be zero. This is the main difference in training algorithms between previous studies [20] and our work.

$$\frac{\partial t_j^l}{\partial V_{mem,j}^l} = \begin{cases} 0 & (if\ t_j^l = 0\ or\ T_{max}) \\ -1 & (o.w.). \end{cases} \quad (8)$$

The weights are updated by eq. (5) and (8) as follows:

$$\Delta w_{ij}^{l-1} = \eta \frac{\partial L}{\partial w_{ij}^{l-1}} = \eta \frac{\partial L}{\partial t_j^l} \frac{\partial t_j^l}{\partial V_{mem,j}^l} \frac{\partial V_{mem,j}^l}{\partial w_{ij}^{l-1}} = \eta \delta_j^l \times \begin{cases} 0 & (if\ t_j^l = 0\ or\ T_{max}) \\ -1 & (o.w.) \end{cases} \times S_i^{l-1}(t_j^l) \quad (9)$$

where $\eta$ is a learning rate. Additionally, the delta values ($\delta_j^l$) are calculated as the weighted sum of the delta values of neurons in the following layer ($\delta_k^{l+1}$). However, only the post-neurons fired by pre-neuron are considered in the calculation. In other words, only the post-neurons which fire later than pre-neurons ($t_k^{l+1} \geq t_j^l$ i.e. $S_j^l(t_k^{l+1}) = 1$) are included in the calculation of delta values of pre-neurons.

$$\delta_j^l = \sum_{k=1}^{N^{l+1}} \delta_k^{l+1} \times w_{jk}^l \ [t_k^{l+1} \geq t_j^l]. \tag{10}$$

Before updating weights, we normalized delta values in each layer to prevent from vanishing gradients in deep layers.

We also set the target firing time of output neurons based on the previous work [20]. We defined $\tau$ as the minimum value of the firing time among the output neurons. Correct output neuron is encouraged to fire first among the output neurons at time $\tau$, and output neurons fired wrongly around $\tau$ have a higher risk of responding incorrectly, so that penalizing as $\alpha_{penalty}$. Then, we set the target firing time of $k$th output neurons as:

$$T_{target,k} = \begin{cases} \tau & : if \ k = answer \\ \tau + \alpha_{penalty} & : if \ k \neq answer, \ t_k^o \leq (T_{max} - \alpha_{penalty}) \\ t_k^o & : if \ k \neq answer, \ t_k^o > (T_{max} - \alpha_{penalty}) \end{cases} \tag{11}$$

## 2.2 NOR-type Synaptic Device Having Asymmetric Floating Gate

On the other hand, various types of emerging memory are being reported as candidates for artificial synaptic devices, which is a key element for configuring SNN in hardware. In our previous work, NOR-type flash memory device was fabricated using a conventional CMOS process [25]. As shown in Fig. 2 (a), this TFT type device has a poly channel and a half-covered poly-Si floating gate (FG) that functions as a charge storage layer. The thickness of blocking $SiO_2$, FG, and tunneling $SiO_2$ are 15 nm, 80 nm, and 7 nm, respectively, and the channel length (the length between source and drain) and the width are 0.5 μm each. Input pulses are presented to each gate (WL), and the currents of the synaptic devices are summed in the common drain line (BL). Hence, the output of vector-by-matrix multiplication (VMM) can be expressed as the current of each post-neuron. In addition, since the current is controlled by three terminals in this FET-type synaptic device, it is more resistant to sneak path issues [26,27] or off-current issues [28,29] than two-terminal devices such as RRAM.

Fig. 2 (b) provides the measured $I_D$-$V_G$ curves of the NOR-type flash memory device. 50 repeated erase pulses ($V_{WL} = -3$ V, $V_{SL} = 5$ V, duration = 100 μs) are applied, and the threshold voltage of the device decreases. The inset of Fig. 2 (b) shows the change of conductance when the device is under the read conditions. The behavior of these synaptic devices is similar to the long-term potentiation (LTP) of the synapse in the nervous systems.

## 3    Results

In this section, we designed a 2-layer fully-connected SNN consisting of NOR-type synaptic devices and neuron circuits. Then, the performance of the network is verified in both the system-level and the circuit-level simulations.

### 3.1    System-Level Simulation

#### 3.1.1 Performance of SNN on MNIST

The system-level simulation is performed on the MNIST datasets to evaluate the performance of SNN based on the NOR-type synaptic devices. To reduce the size of the network, the edge of the image is removed and resized to 20 × 20. The parameters used in the simulation are shown in Table 1. The input data is transformed into a TTFS spike train over 64 time steps, and the $\alpha_{penalty}$ mentioned in section 2.1.3 is set to 1. The batch size of training is 1, and the initial learning rate is set to 0.02, but it gradually drops as training continues. The threshold of neurons in all layers is 1.6 V, and the winner neuron of SNN is determined as the neuron that fires first among neurons in the last readout-layer. However, if any output neuron does not spike until the last time step, it is evaluated by considering the membrane voltage of the output neuron at the last time step. The weights are initialized using the initialization method proposed by K. He [30]. The initial weight distribution is given by:

$$W^l \sim N\left(I_{init}, \frac{2}{n_{in}^l}\right) \qquad (12)$$

where $n_{in}^l$ represents the number of input nodes in the $l$th layer. However, by changing the mean of the normal distribution to a positive value ($I_{init}$) rather than 0, many hidden neurons are fired at the start of training, leading to participation in data propagation.

Weights trained by the off-chip learning rules described in section 2.1 are transferred to the HNN. Prior to being transferred, the weights are normalized and quantized to the 101 states. The weight for one synapse in the SNN is represented by the conductance difference between two synaptic devices as follows:

$$w_{ij}^l = G_{ij}^{l(+)} - G_{ij}^{l(-)} \qquad (13)$$

where $G_{ij}^l$ is one of the measured conductance value of the asymmetric FG synaptic device ($G(1), G(2)$, .. $G(50)$). For all the positive $w_{ij}^l$, $G_{ij}^{l(-)}$ is set to $G(1)$ ($G_{min}$), and conversely, for all the negative $w_{ij}^l$, $G_{ij}^{l(+)}$ is set to $G(1)$ ($G_{min}$). If the $w_{ij}^l$ is 0, the conductance of both synaptic devices is set to $G(1)$ ($G_{min}$) [31]. The quantized target weights of each synapse can be transferred by applying the corresponding index number of pulses to devices in the synaptic array [32].

Fig. 3 (a) presents MNIST accuracy as a parameter of the width of the hidden layer in 2-layer SNN. The accuracy for 60,000 training sets and 10,000 test sets not used for training is indicated by dotted and solid lines, respectively. As the number of hidden neurons increased, it is observed that the accuracy increased. The accuracy of the network (400 - 512 - 10) is 99.21% for the training set, and 96.90% for the test set. The accuracy shows the degradation of about 1% compared to those of rate-encoded networks of similar size [33,34]. Fig. 3 (b) shows the recognition accuracy with the total number of time steps per image. The time steps, representing the resolution of the image, can be reduced to 8 without significant degradation of the accuracy (0.15% degradation for 512 hidden neurons).

### 3.1.2 Effects on Accuracy by Variation in Hardware Implementation

Changes in device characteristics caused by process variations during manufacturing negatively affect the operation of synaptic devices and neuronal circuits, which reduces the recognition accuracy of the SNN implemented in hardware. Several types of variation have been analysed in previous studies [35-40]. We classify the four major variations as follows:

1) device-to-device variation in the synaptic array [35,36],

2) firing threshold variation in the neuron circuits [37],

3) stuck-at-off variation in the synaptic array [38-40], and

4) stuck-at-off variation in the neuron circuits.

Note that our network does not take into account the pulse-to-pulse variation considered in many previous studies [24,41] since the weights obtained through off-chip learning are transferred once to the synaptic devices in the array. The recognition accuracy with the variation of device characteristics is compared with that of conventional rate-encoded networks with the same size. In the rate-encoded network, the number of input spikes follows a Poisson distribution, and the weights are also quantized with the same resolution in the same manner as in the TTFS network.

We first mathematically model the device-to-device variation in the synaptic array as follows:

$$W^l \leftarrow W^l \times N(1, \sigma^2_{weight}) \tag{14}$$

where $W^l$ denotes the overall quantized weights, and $N(1, \sigma^2_{weight})$ stands for the normal distribution with mean 1 and standard deviation $\sigma_{weight}$. Also, weights with a value of 0 are set to random numbers with normal distribution N(0, $\sigma_{weight}$). In addition, variation of neuron thresholds is also modeled by normal distribution:

$$V^l_{th} \leftarrow \max(V^l_{th} \times N(1, \sigma^2_{th}), 0). \tag{15}$$

However, negative neuron thresholds are very difficult to implement in hardware, so they follow a clipped normal distribution. Lastly, in large-sized synapse and neuron arrays, a stuck-at-off fault where one of the devices is not working should be considered. Dead synaptic devices can cause the current to not flow even the input pulse is applied, which results in a fatal error in the weighted sum. Further, if the neuron block dies, there may be cases where the current cannot be integrated into the capacitor, or the spike cannot be emitted even if the membrane voltage exceeds the neuron threshold. We defined the stuck-at-off ratio as the number of dormant synaptic devices (neurons) relative to the total number of devices (neurons) in the array and named it $R_{synapse}$ ($R_{neuron}$). The conductance of the dead synapse is assumed to 0, and the input by a dead neuron is assumed to 0 regardless of the membrane voltage.

Fig. 4 (a) - (d) show the degradation of the accuracy as the variation in the synaptic array gets worse. Compared to the rate-encoded network, TTFS network is vulnerable to the variation since a single neuron can contribute only one spike in the inference process. In TTFS network, even if only one spike disappears (or even one false spike occurs), it makes a big error in the overall weighted sum. In particular, it is observed that the network with a small number of hidden neurons shows severe degradation of accuracy due to the importance of each neuron. Therefore, synaptic arrays of TTFS networks should be finely controlled so that the variation to be minimized. For example, device-to-device variation can be reduced by precision tuning using the read-write-verify scheme in the weight transfer process [42].

## 3.2 Circuit Level Simulation

In the previous section, we verified that data encoded by TTFS method can be effectively trained through temporal backpropagation. However, non-ideal issues that have not been considered at system-level simulations, such as parasitic resistance or capacitance, can be a problem in large-sized arrays. In this section, we propose the neuron blocks suitable for TTFS encoded SNN, and simulate fully connected HNN at circuit-level using SPICE. Through this, we will investigate the result of the system-level simulation is reasonable, and how much power the full network consumes in the inference process.

### 3.2.1 SNN Architecture

Fig. 5 (a) depicts a schematic diagram of fully connected HNN. The entire network is composed of the synaptic array and neuron circuits. The NOR-type synaptic device is modeled with a voltage-controlled current source (VCCS) behavioral model with 3 terminals (Gate, Drain, Source) as shown in Fig. 5 (b) [43,44].

The current between the drain and the source is determined by the voltage difference between the gate and the source. As shown in Fig. 2 (b), the behavior modeling of the synaptic device was based on the results measured by increasing the gate voltage from 0 V to 3 V in 60 mV steps.

In addition, neuron circuits consist of a current mirror, an I&F block, and a refractory period generator. Fig. 5 (b) depicts a modeled synaptic array and a current mirror designed for summing and subtracting currents. A single wordline (WL) corresponds to an input, and the weighted sum of 400 inputs is expressed as the sum of current flowing through the bitline (BL). Currents flowing in the positive and negative synaptic arrays are copied through the respective current mirrors so that the net charge is integrated in the membrane capacitor ($C_{mem}$). Before the input pulse is presented, $V_{mem}$ of all neurons are initialized to $V_{DD2}$ by $V_{init}$. If the initial membrane voltage of the neuron is not $V_{DD2}$ (e.g. 0 V), the negative charge created by the inputs in the early stage of the time domain cannot be integrated into the capacitor, so the final result of the weighted sum can be distorted. Since SNN encoded by TTFS method assumes that one neuron spikes at most once, the neurons already fired should enter the refractory period so that no more spikes are generated. To implement this, the output neuron that has already fired keeps $V_{refrac}$ in a high state until the corresponding input ends. This causes M7 and M8 to turn off so that no more current flows through the synaptic array.

Fig. 5 (c) shows the circuit of the refractory period generator (RPG). The block for generating $V_{refrac}$ is based on the structure of the latch. Before the input pulse is presented, M12 is turned on by $V_{init}$, so the output node of RPG is initialized to the ground state. Then, as soon as the I&F block fires, M10 and M11 turn on and $V_{refrac}$ goes to high state, which is maintained until a new input data is presented.

Fig. 5 (d) represents the I&F block constituting the neuron [45]. If the membrane voltage exceeds the $V_{th}$ of M14 by the integrated charge, node 1 in the high state changes to the low state. This brings node 2 to the high state. After the delay time by $C_{pulse}$, the voltage at node 2 turns M21 on and puts node 1 back in high state, and returns node 2 to original state. The W/L ratio of M16 acting as a resistor and the value of $C_{pulse}$ determine the width of a spike generated in the output node. In addition, the W/L ratio of M14 and M21 determines the voltage of node 1, so it affects the threshold of the neuron. After I&F block fires, $V_{refrac}$ turns M13 on to keep $V_{mem}$ as the ground state.

### 3.2.2 Performance in Circuit-Level Simulation

Among the networks simulated in the previous section 3.1, a relatively light network, the 400-128-10 sized network is simulated using a circuit simulator (HSPICE) with a predictive technology models (PTMs). Circuit-level simulation was performed with 0.35 μm CMOS technology, and the parameters of the components in circuits are shown in Table 2.

Fig. 6 provides the waveforms of some nodes in the process of inferencing MNIST data sets. Before the input pulses are presented, $V_{init}$ is first presented to initialize the membrane capacitors in I&F block and RPG block. After that, as shown in Fig. 6 (a), all inputs are transformed into time-to-first spike pulses with a duration of 0.5 μs over 8 time steps. The interval between each time step of the input pulse is also set to 0.5 μs. The rising and falling times of input pulses are each set to 0.1 μs. Fig. 6 (b) shows transient waveforms of some nodes in hidden neurons. The currents flowing through the synaptic array by the input pulses are integrated into the capacitor of the hidden neurons, and when $V_{mem}$ exceeds the threshold of the neuron, corresponding neuron fires and presents a spike with a width of 0.5 μs to the post-layer. At the very moment the neuron fires, $V_{refrac}$ generated by each RPG prevents further integration of charge into the fired neuron. Finally, Fig. 6 (c) represents the membrane voltage of neurons in the output layer. As in the system simulation, the class of the earliest fired output neuron is the result predicted by SNN. However, in some rare cases when no output neuron fires, the neuron with the highest membrane voltage is considered as the winner neuron.

Fig. 7 compares the results of system-level and circuit-level simulations. The size of the network is 400 - 128 - 10, and the number of time step for each image is set to 8. Fig. 7 (a)-(c) shows the firing times of input, hidden and output layers in the system-level simulation of one image. The *x*-axis of the three raster plots represents the time and the *y*-axis stands for the index number of neurons in each layer. Fig. 7 (e) and (f) depict a raster plot of spike timing in hidden neurons and $V_{mem}$ of output neurons in the circuit-level simulation for the same image. By comparing the firing times of hidden neuron and output neuron shown in (b), (e) and (c), (f), it is observed that the results of both simulations are similar.

In addition, we also simulated the circuits for 1000 randomly selected MNIST data sets. Fig. 7 (d) shows the result of comparing the firing time of the winner neuron obtained by simulations at system-level (*x*-axis) and circuit-level (*y*-axis). Since the system-level simulation was performed during 8 discrete time steps, the firing time in the system-level is a discrete value. The firing times of the two simulations are not perfectly matched, but they show almost the same tendency, which means the proposed SNN shown at the circuit level works pretty much like that at the system level. Indeed, the proposed SNN has reached 94.9% accuracy for networks having 128 hidden neurons at the circuit-level. This accuracy is only 0.1% lower than the 95.0% accuracy in a system level simulation. The reason for the slight decrease in accuracy is that the off current in the synaptic array is not considered at the system-level. Also, calculating the weighted sum through discrete time steps in system-level simulation can cause a difference from actual circuit operation.

### 3.2.3 Power Measurements

In this section, we estimate the power consumed by the TTFS network at the circuit-level and compare the results with that of the rate-encoded network. The biggest advantage of TTFS encoding method is that it requires much fewer pulses compared to the conventional rate encoding method, which results in lower power consumption. TTFS and rate-encoded networks, each with the same number (128) of hidden neurons are simulated for 100 randomly picked MNIST data sets, and the total time steps for each image are set to 8.

Fig. 8 shows the amount of power consumed by each block in the proposed SNN. As categorized in section 3.2.1, the entire network consists of synapse array (SA) and neuron circuits, and specifically, the neuron is composed of a current mirror (CM), a circuit for integrate and fire (IF), and a refractory period generator (RPG). Fig. 8 (a) depicts the power consumed in the inference process of TTFS network. The entire network consumes 353.6 μW, and it is observed that most (~ 90%) of the power is consumed by the components in the 1st layer. In particular, I&F block accounts for a remarkable proportion of power consumption. This is because not only power is consumed to generate the pulse, but also subthreshold leakage current flows due to the membrane voltage of the neuron below $V_{th}^l$. In the I&F block depicted in Fig. 5 (d), even if $V_{mem}$ does not exceed $V_{th}$, M14 can be finely turned on if $V_{mem}$ is a positive value. This creates a leakage path through M14 and M15, allowing current to flow even the neuron is not fired. Since the number of spikes in TTFS network is small, this standby power occupies a relatively large portion as much as the power required to generate spikes. Improving the structure of I&F circuits to deal with this issue can be a topic for further study.

Fig. 8 (b) represents the power consumed in the rate-encoded networks. In the circuit-level simulation, each input spike of the rate-encoded network is filled from the last time steps [46]. Compared to the TTFS encoding method, the rate-encoding method requires much more pulses to represent an image, so the currents in the synapse array and current mirror are very large. Likewise, the number of spikes generating in each layer is greater, so the power consumed by I&F block is larger. Unlike TTFS network, the rate-encoded network does not require a refractory period generator, but the power that can be saved is very small (~2%). It is obtained that the entire network consumes 1240 μW of power, which is 3.5 times more than that of the TTFS network.

The power consumption ratio of the two networks increases as the total time steps per each image increases. Fig. 9 (a) shows the required number of spikes and consumed energy as a function of time step. The solid line represents the average value of the spike numbers required to compute an image at the system-level. The required spike number is the sum of spikes generated in all layers. The number of pulses in the TTFS network is only counted until the winner neuron of the output layer fires, and the number in the rate-encoded network is counted until the final time step. When the input is converted to a spike rate, the number of spikes required to express the same values increases as the time step increases. On average, if the total time steps are 4, only 49.5 spikes are required, whereas 30793 spikes are needed when the time step reaches 256. On the other hand, the number of spikes in TTFS encoding method is nearly constant at about 162 regardless of the total time steps. Therefore, as the resolution of input data increases, the difference between the required spike numbers of the two networks increases.

Meanwhile, the dashed lines represent the average energy required to compute an image as a result of circuit-level simulation. Since the time required to compute the image depends on the time step, the energy is compared between two networks. On average, the rate-encoded network consumes 5.65 nJ of energy per image at a time step number of 4 and 372 nJ at a time step number of 256. On the other hand, TTFS network consumes 2.16 nJ at a time step number of 4 and 24.6 nJ at a time step number of 256 to compute one image. As the total time step of TTFS network grows, the number of spikes is not changed, but the consumed energy is increased. This is because the amount of energy consumed by the leakage path in I&F block is proportional to the time for processing an image. Rate-encoded networks are also affected by this leakage, but the relative proportion of the leakage in total energy consumption is less than that in TTFS network due to a large number of spikes. Hence, the consumed energy of the rate encoded network is almost proportional to the required spike number. Meanwhile, TTFS, network uses a small number of spikes, tends to increase the consumed energy even if the required spike number is constant. Nevertheless, the superiority of TTFS network in terms of power-efficiency is increased as the time step increases compared to the rate encoded network. Finally, the ratio of power efficiency of TTFS networks to rate-encoded networks reaches 15.75 at a time step number of 256.

The TTFS network also has an advantage in terms of the latency, the time it takes to infer the answer. Fig. 9 (b) shows the latency of the TTFS network and rate-encoded network as a parameter of the number of total time steps. The latency of the TTFS network was calculated as the average value of the time until the emission of the first spike at the output layer. On the other hand, the rate-encoded network can only make a decision at the last time step, so the latency of the network is equal to the value of the total time step, as indicated by the dashed line. It is observed that the TTFS network with 128 hidden neurons can make a decision about 5 times faster than rate-encoded network of the same size.

## 4  Conclusion

In this study, we have evaluated the performance of the SNN consisting of NOR-type asymmetric FG synaptic devices and neuron circuits at system-level and circuit-level. Input data was encoded as the time of the input spikes (time-to-first spike: TTFS), and the network was trained by temporal backpropagation which is a learning method suitable for networks applying TTFS encoding method. The neural network with 512 hidden neurons showed a competitive accuracy of 96.90 % for the cropped MNIST data sets. We also investigated the impact of the non-ideal characteristics of the synaptic array and neuron circuits on accuracy. These results can be a guideline that informs which level of the variation is allowed in TTFS network. In addition, we proposed a neuron circuit for inferencing temporal data and modeled the synapse device to demonstrate the operation of the full network. Simulating an SNN with 128 hidden neurons in SPICE gives 94.9% accuracy for 1000 MNIST data sets, which is almost no degradation compared to the system-level simulation. We also analyzed the power consumed in the inference process by each block in SNN. When using 8 time steps in a 400-128-10 size network, the TTFS network showed approximately 3.5 times higher power efficiency compared to the rate-encoded network. At the same network size, TTFS networks showed significantly lower energy consumption and shorter latency than rate-encoded networks. The difference in energy consumption between the two networks increases as the number of time step increases.

## 5  Acknowledgements

This work was supported by the Brain Korea 21 Plus Project in 2020, and MOTE (Ministry of Trade, Industry & Energy (10080583) and KSRC (Korea Semiconductor Research Consortium) support program for the development of the future semiconductor device.

**Compliance with ethical standards**

**Conflict of Interest**: The authors declare that they have no conflict of interest.

**References**

[1] A. Krizhevsky, I. Sutskever, and G. E. Hinton, "Imagenet classification with deep convolutional neural networks," *Adv. Neural. Inf. Process. Syst. (NIPS)*, pp. 1097-1105, 2012. http://papers.nips.cc/paper/4824-imagenet-classification-with-deep-convolutional-neural-networks

[2] Y. LeCun, Y. Bengio, and G. Hinton, "Deep learning," *Nature*, vol. 521, pp. 436-444, 2015. https://doi.org/10.1038/nature14539

[3] T. Gokmen, and Y. Vlasov, "Acceleration of deep neural network training with resistive cross-point devices: Design considerations," *Front. Neurosci.*, vol. 10, no. 333, pp. 1-13, 2016. https://doi.org/10.3389/fnins.2016.00333

[4] S. Ambrogio, P. Narayana, H. Tsai, R. M. Shelby, I. Boybat, C. Nolfo, S. Sidler, M. Giordano, M. Bodini, N. C. P. Farinha, B. Kileen, C. Cheng, Y. Jaoudi, G. W. Burr, "Equivalent-accuracy accelerated neural-network training using analogue memory," *Nature*, vol. 558, pp. 60-67, https://doi.org/10.1038/s41586-018-0180-5

[5] R. Collobert, J. Weston, "A unified architecture for natural language processing: Deep neural networks with multitask learning," *Proc. Int. Conf. Mach. Learn. (ICML)*, pp. 160-167, 2008. https://dl.acm.org/doi/10.1145/1390156.1390177

[6] D. E. Rumelhart, G. E. Hinton, and R. J. Williams, "Learning representations by back-propagating errors," *Nature*, vol. 323, pp. 533-536, 1986. https://doi.org/10.1038/323533a0

[7] J. Ba, R. Caruana, "Do deep nets really need to be deep?," *Adv. Neural. Inf. Process. Syst. (NIPS)*, pp. 2654-2662, 2014. http://papers.nips.cc/paper/5484-do-deep-nets-really-need-to-be-deep

[8] A. Taherkhani, A. Belatreche, Y. Li, G. Cosma, L. P. Maguire, and T. M. McGinnity, "A review of learning in biologically plausible spiking neural networks." *Neural Netw.*, vol. 122, pp. 253-272, 2020. https://doi.org/10.1016/j.neunet.2019.09.036

[9] JL. McClelland, and D. E. Rumelhart, "Parallel distributed processing," *Explorations in the microstructure of cognition 2*, pp. 216-271, 1986.

[10] M. Pfeiffer, and T. Pfeil, "Deep learning with spiking neurons: opportunities and challenges." *Front. Neurosci.*, vol. 12, no. 774, 2018. https://doi.org/10.3389/fnins.2018.00774

[11] B. Rueckauer, I. Lungu, Y. Hu, M. Pfeiffer, and S. C. Liu, "Conversion of continuous-valued deep networks to efficient event-driven networks for image classification," *Front. Neurosci.*, vol. 11, no. 682, 2017. https://doi.org/10.3389/fnins.2017.00682

[12] K. He, X. Zhang, S. Ren, and J. Sun, "Deep residual learning for image recognition." *Proc. IEEE Comput. Soc. Conf. Comput. Vis. Pattern. Recognit. (CVPR)*, pp. 770-778, 2016. http://openaccess.thecvf.com/content_cvpr_2016/html/He_Deep_Residual_Learning_CVPR_2016_paper

[13] E. Hunsberger, C. Eliasmith, "Training spiking deep networks for neuromorphic hardware," *arXiv:*1611.05141, 2016. https://arxiv.org/abs/1611.05141

[14] W. Maass, and T. Natschläger, "Emulation of Hopfield networks with spiking neurons in temporal coding." *J. Comput. Neurosci.*, Springer, pp. 221-226, 1998. https://doi.org/10.1007/978-1-4615-4831-7_37

[15] S.M. Bohte, J.N. Kok, and H.L. Poutre, "Error-backpropagation in temporally encoded networks of spiking neurons," *Neurocomputing*, vol. 48, no. 1-4, pp. 17-37, 2002. https://doi.org/10.1016/S0925-2312(01)00658-0

[16] Q. Yu, H. Tang, K. C. Tan, and H. Yu, "A brain-inspired spiking neural network model with temporal encoding and learning," *Neurocomputing*, vol. 138, pp. 3-13, 2014. https://doi.org/10.1016/j.neucom.2013.06.052

[17] H. Mostafa, "Supervised learning based on temporal coding in spiking neural networks," *IEEE Trans. Neural. Netw. Learn. Syst.*, vol. 29, no. 7, pp. 3227-3235, 2017. https://doi.org/10.1109/TNNLS.2017.2726060

[18] I. M. Comsa, K. Potempa, L. Versari, T. Fischbacher, A. Gesmundo, and J. Alakuijala, "Temporal coding in spiking neural networks with alpha synaptic function," *Proc. IEEE Int. Conf. Acoust. Speech Signal Process (ICASSP)*, pp. 8529 - 8533, 2020. https://doi.org/10.1109/ICASSP40776.2020.9053856

[19] B. Rueckauer, and S. C. Liu, "Conversion of analog to spiking neural networks using sparse temporal coding." *IEEE Int. Symp. Circuits. Syst. Proc (ISCAS).*, pp. 1-5, 2018. https://doi.org/10.1109/ISCAS.2018.8351295

[20] S. R. Kheradpisheh, and T. Masquelier, "Temporal backpropagation for spiking neural networks with one spike per neuron," *Int. J. Neural. Syst.*, 2020. https://doi.org/10.1142/S0129065720500276

[21] R. Vaila, J. Chiasson, and V. Saxena, "A Deep Unsupervised Feature Learning Spiking Neural Network with Binarized Classification Layers for EMNIST Classification," *arXiv:*2002.11843, 2020. https://arxiv.org/abs/2002.11843

[22] C. Lee, G. Srinivasan, P. Panda, and K. Roy, "Deep spiking convolutional neural network trained with unsupervised spike-timing-dependent plasticity," *IEEE Trans. Cogn. Dev. Syst.*, vol. 11, no. 3, pp. 384-394, 2018. https://doi.org/10.1109/TCDS.2018.2833071

[23] S. R. Kheradpisheh, M. Ganjtabesh, S. J. Thorpe, and T. Masquelier, "STDP-based spiking deep convolutional neural networks for object recognition," *Neural Netw.*, vol. 99, pp. 56-67, 2018. https://doi.org/10.1016/j.neunet.2017.12.005

[24] S. Oh, C. H. Kim, S. Lee, J. Kim, and J. H. Lee, "Unsupervised online learning of temporal information in spiking neural network using thin-film transistor-type NOR flash memory devices," *Nanotechnology*, vol. 30, no. 435206, 2019. https://doi.org/10.1088/1361-6528/ab34da.

[25] C. H. Kim, S. Lee, S. Y. Woo, W. M. Kang, S. Lim, J. H. Bae, J. Kim, and J. H. Lee, "Demonstration of unsupervised learning with spike-timing-dependent plasticity using a TFT-type NOR flash memory array," *IEEE Trans. Electron Devices*, vol. 65, no. 5, pp. 1774-1780, 2018. https://doi.org/10.1109/TED.2018.2817266

[26] E. Linn, R. Rosezin, C. Kugeler, and R. Waser, "Complementary resistive switches for passive nanocrossbar memories," *Nat. Mater.*, vol. 9, no. 5, pp. 403-406, 2010. https://doi.org/10.1038/NMAT2748

[27] J. Liang, and H-S. P. Wong, "Cross-point memory array without cell selectors—Device characteristics and data storage pattern dependencies," *IEEE Trans. Electron Devices*, vol. 57, no. 10, pp. 2531-2538, 2010. https://doi.org/10.1109/TED.2010.2062187

[28] P.F. Chiu, and B. Nikolic, "A differential 2R crosspoint RRAM array with zero standby current," *IEEE Trans. Circuits Syst. II Express Briefs*, vol. 62, no. 5, pp. 461-465, 2014. https://doi.org/10.1109/TCSII.2014.2385431

[29] S. Kim, X. Liu, J. Park, S. Jung, W. Lee, J. Woo, J. Shin, G. Choi, C. Cho, S. Park, D. Lee, E. J. Cha, B. H. Lee, H. D. Lee, S. G. Kim, S. Chung, and H. Hwang, "Ultrathin (<10nm) $Nb_2O_5/NbO_2$ Hybrid Memory with Both Memory and Selector Characteristics for High Density 3D Vertically Stackable RRAM Applications," *Symp. VLSI Tech.*, 2012. https://doi.org/10.1109/VLSIT.2012.6242508

[30] K. He, X. Zhang, S. Ren, and J. Sun, "Delving Deep into Rectifiers: Surpassing Human-Level Performance on ImageNet Classification," *IEEE Int. Conf. Comput. Vis. Workshops (ICCV)*, pp. 1026-1034, 2015. https://www.cv-foundation.org/openaccess/content_iccv_2015/html/He_Delving_Deep_into_ICCV_2015_paper

[31] G. W. Burr, R. M. Shelby, S. Sidler, C. di. Norfo, J. Jang, B. Irem, R. S. Shenoy, P. Narayanan, K. Virwani, E. U. Giacometti, B. N. Kurdi, and H. Hwang, "Experimental Demonstration and Tolerancing of a Large-Scale Neural Network (165000 Synapses) Using Phase-Change Memory as the Synaptic Weight Element," *IEEE Trans. Electron Devices*, Vol. 62, No, 11, pp. 3498-3507, 2015. https://www.doi.org/10.1109/ted.2015.2439635.

[32] S. Lim, D. Kwon, J. H. Eum, S. T. Lee, J. H. Bae, H. Kim, C. H. Kim, B. G. Park, and J. H. Lee, "Highly Reliable Inference System of Neural Networks Using Gated Schottky Diodes," *IEEE Journal of the Electron Devices Society*, vol. 7, pp. 522-528, 2019. https://doi.org/10.1109/JEDS.2019.2913146

[33] D. Kwon, S. Lim, J. H. Bae, S. T. Lee, H. Kim, C. H. Kim, B. G. Park, and J. H. Lee, "Adaptive weight quantization method for nonlinear synaptic devices," *IEEE Trans. Electron Devices*, vol. 66, no. 1, pp. 395-401, 2018. https://doi.org/10.1109/TED.2018.2879821

[34] E. O. Neftci, C. Augustine, S. Paul, and G. Detorakis, "Event-Driven Random Back-Propagation: Enabling Neuromorphic Deep Learning Machines," *Front. Neurosci.*, vol. 11, no. 324, 2017. https://doi.org/10.3389/fnins.2017.00324

[35] S. Yu, B. Gao, Z. Fang, H. Yu, J. Kang, and H.-S. P. Wong, "A low energy oxide-based electronic synaptic device for neuromorphic visual systems with tolerance to device variation," *Adv. Mater.*, vol. 25, no.12, pp. 1774-1779, 2013. https://doi.org/10.1002/adma.201203680

[36] D. Querlioz, O. Bichler, P. Dollfus, and C. Gamrat, "Immunity to device variations in a spiking neural network with memristive nanodevices," *IEEE Trans. Nanotechnol.*, vol. 12, no. 3, pp. 288-295, 2013. https://doi.org/10.1109/TNANO.2013.2250995

[37] Y. Sakemi, K. Morino, T. Morie, and K. Aihara, "A Supervised Learning Algorithm for Multilayer Spiking Neural Networks Based on Temporal Coding Toward Energy-Efficient VLSI Processor Design," *arXiv*:2001.05348, 2020. https://arxiv.org/abs/2001.05348.

[38] Y. Li, Z. Wang, R. Midya, Q. Xia, and J. J. Yang, "Review of memristor devices in neuromorphic computing: materials sciences and device challenges," *J. Phys. D. Appl. Phys.*, vol. 51, no. 50, 2018. https://doi.org/10.1088/1361-6463/aade3f

[39] D. Kwon, S. T. Lee, H. Kim, J. H. Bae, S. Lim, K. Yeom, J. Kim, B. G. Park, and J. H. Lee, "On-Chip Training Spiking Neural Networks Using Approximated Backpropagation with Analog Synaptic Devices," *Front. Neurosci.*, vol. 14, no. 423, 2020.

[40] S. N. Truong, "Single Crossbar Array of Memristors With Bipolar Inputs for Neuromorphic Image Recognition," *IEEE Access*, vol.8, pp. 69327-69332, 2020. https://doi.org/10.1109/ACCESS.2020.2986513

[41] P. Y. Chen, B. Lin, I. T. Wang, T. H. Hou, J. Ye, S. Vrudhula, J. S. Seo, Y. Cao, and S. Yu, "Mitigating effects of non-ideal synaptic device characteristics for on-chip learning," *ICCAD IEEE ACM Int. Conf. Comput. Aided. Des.*, 2015. https://doi.org/10.1109/ICCAD.2015.7372570.

[42] F. Alibart, L. Gao, B. D. Hoskins, and D. B. Strukov, "High precision tuning of state for memristive devices by adaptable variation-tolerant algorithm," *Nanotechnology*, vol. 23, no. 075201, 2012. https://doi.org/10.1088/0957-4484/23/7/075201.

[43] M. J. Sharifi, and Y. M. Banadaki, "General SPICE models for memristor and application to circuit simulation of memristor-based synapses and memory cells," *J. Circuit. Syst. Comp.*, vol. 19, no. 2, pp. 407-424, 2010. https://doi.org/10.1142/S0218126610006141

[44] H. Kim, and B. G. Park, "Solving Overlapping Pattern Issues in On-Chip Learning of Bio-Inspired Neuromorphic System with Synaptic Transistors," *Electronics*, vol. 9, no. 13, 2020. https://doi.org/10.3390/electronics9010013

[45] W. M. Kang, C. H. Kim, S. Lee, S. Y. Woo, J. H. Bae, B. G. Park, and J. H. Lee, "A Spiking Neural Network with a Global Self-Controller for Unsupervised Learning Based on Spike-Timing-Dependent Plasticity Using Flash Memory Synaptic Devices," *Proc. Int. Jt. Conf. Neural. Netw. (IJCNN)*, 2019. https://doi.org/10.1109/IJCNN.2019.8851744

[46] S. Hwang, H. Kim, J. Park, M. W. Kwon, M. H. Baek, J. J. Lee, and B. G. Park, "System-level simulation of hardware spiking neural network based on synaptic transistors and I&F neuron circuits," *IEEE Electron Device Lett.*, vol. 39, no. 9, pp. 1441-1444, 2018. https://doi.org/10.1109/LED.2018.2853635.

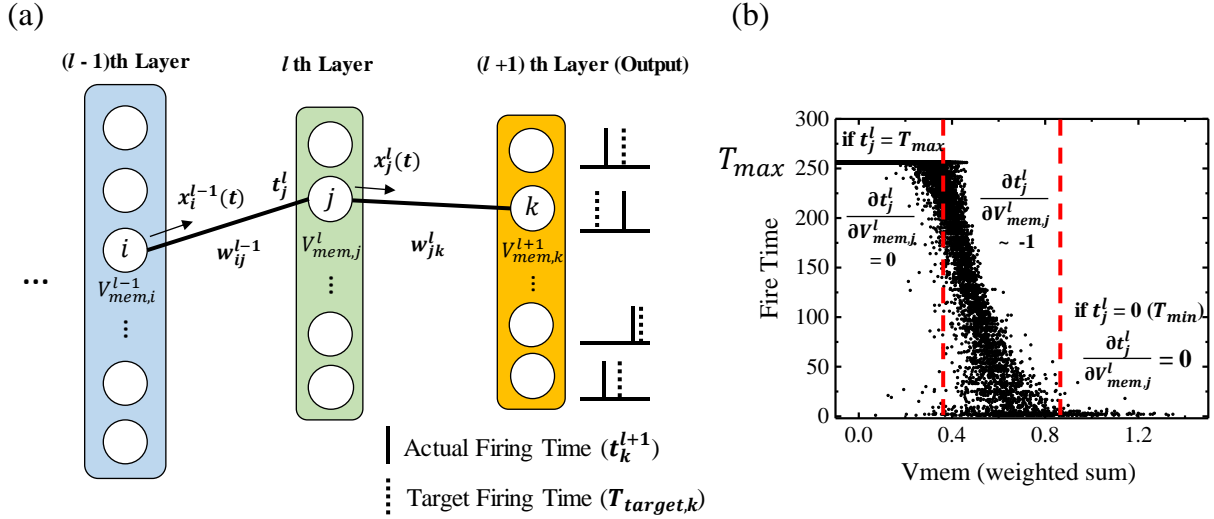

Figure 1. (a) Schematic illustration of a fully connected multi-layer neural network. $t_j^l$ and $V_{mem,j}^l$ represents the firing time and the membrane voltage of $j$th neuron of $l$th layer, respectively. $w_{ij}^{l-1}$ is the weights between $i$th neuron of ($l$-1)th layer and $j$th neuron of $l$th layer. (b) The relationship between the firing time and the membrane potential of the neuron through Monte Carlo simulation. If the firing time of a neuron is not $T_{max}$ or $T_{min}$, the firing time tends to decrease as the membrane voltage (weighted sum) integrated into the neuron increases.

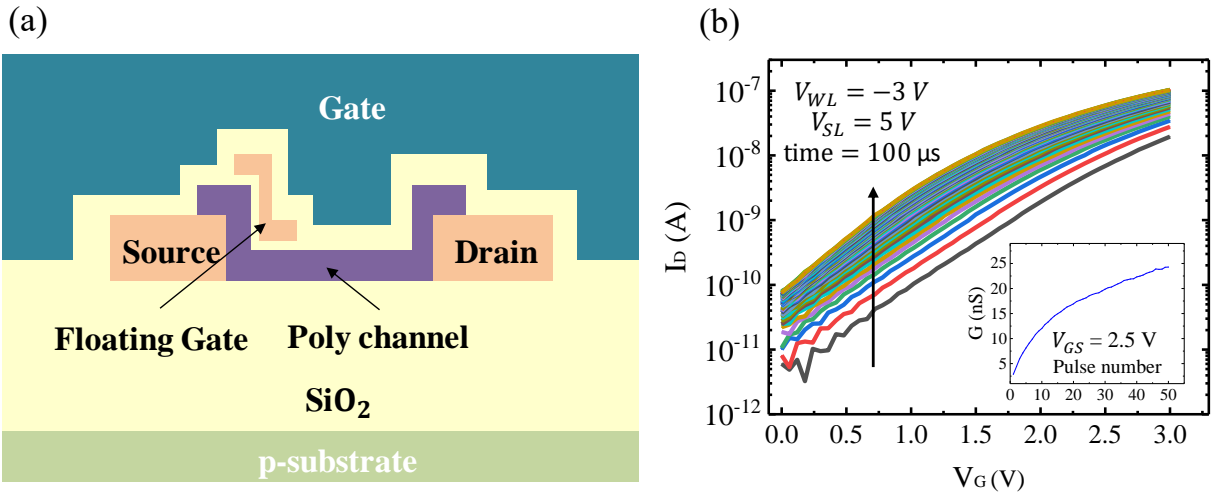

Figure 2. (a) Cross-sectional view of NOR-type flash memory having an asymmetric floating gate. (b) Measurement of $I_D$-$V_G$ characteristics of the NOR-type device changed by applying a consecutive erase pulse. The inset shows the change of conductance in the read condition according to the applied pulse number.

| Parameters | Description | Value |
|---|---|---|
| $T_{max}$ | Time step for each image | 64 |
| $\alpha_{penalty}$ | Penalized term for the incorrect neuron | 1 |
| $\eta$ | Learning rate | 0.02 |
| $V_{th}^l$ | Threshold voltage of I&F neuron in $l$ th layer | 1.6 V |
| $I_{init}$ | Initialization term | 0.1 |

Table 1. Parameters used in the system-level simulation

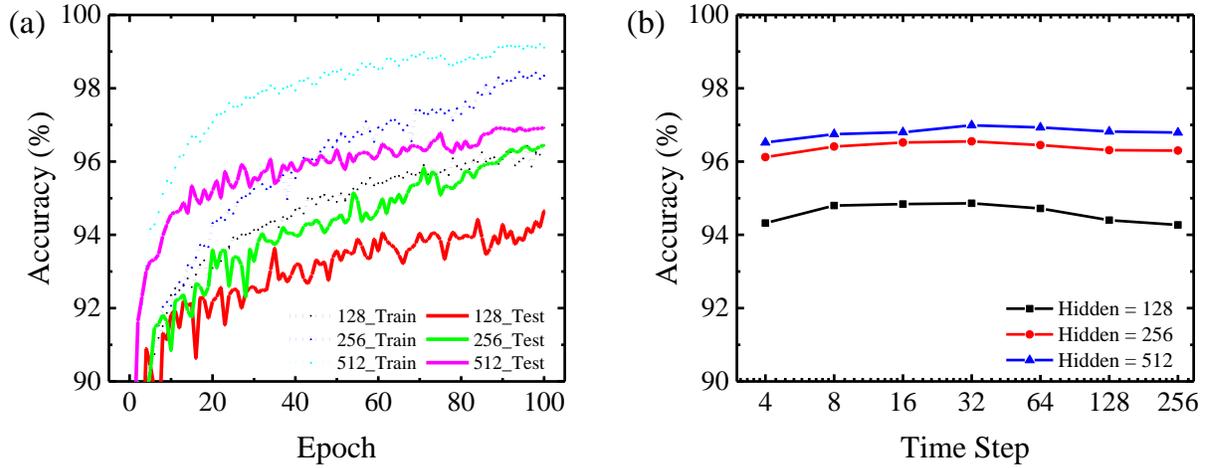

Figure 3. (a) Training curves of the TTFS network as a parameter of the number of neurons in the hidden layer. The dotted line represents the accuracy for training sets, and the solid line represents the accuracy for test sets. (b) The recognition rate of the TTFS network as a parameter of the total time steps.

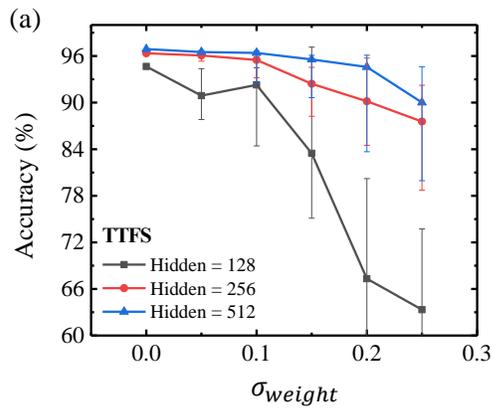
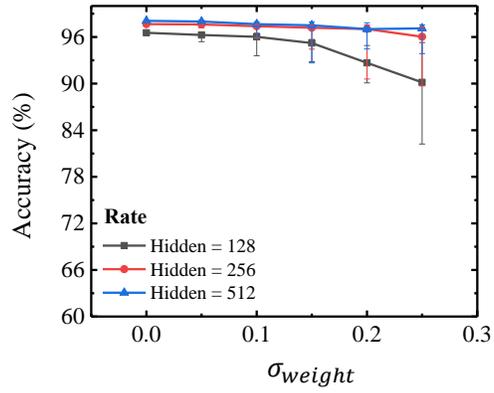

(a)

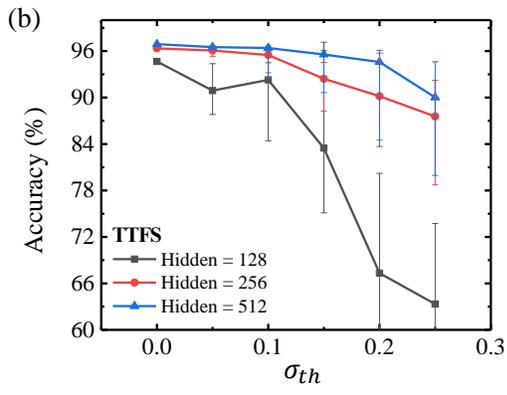
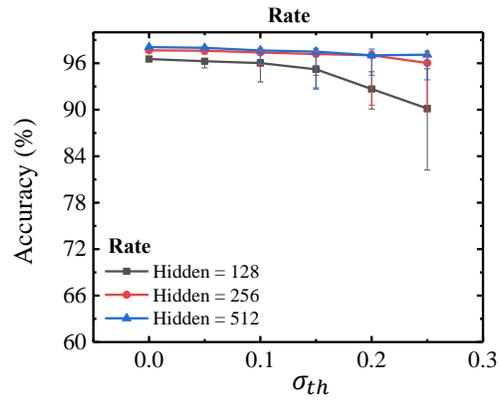

(b)

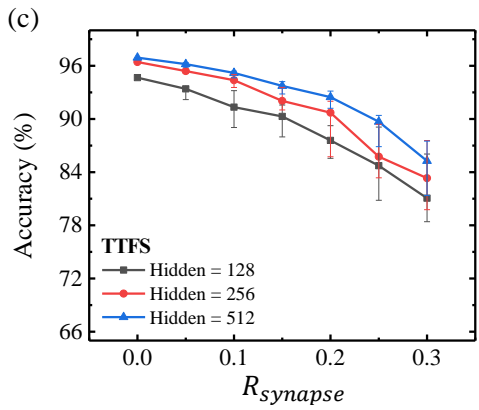
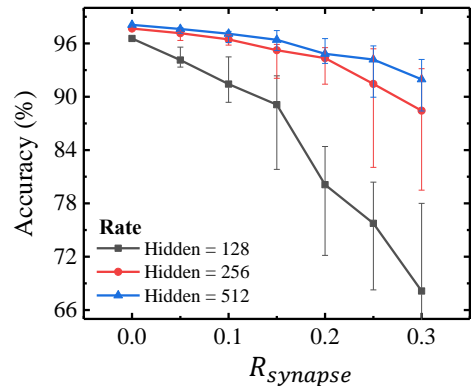

(c)

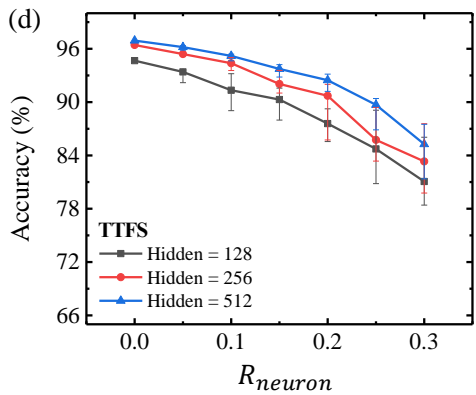
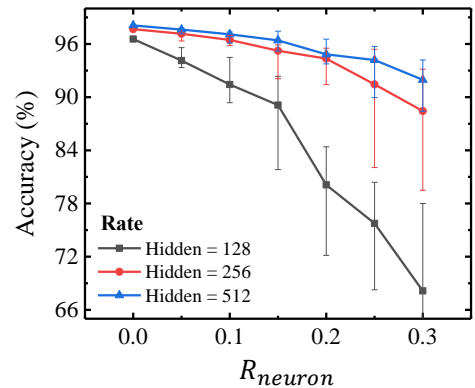

(d)

Figure 4. Evaluation of the (Left) TTFS and (Right) rate-encoded network as a parameter of (a) device-to-device variation in the synaptic array, (b) firing threshold variation in the neuron circuits, (c) the stuck-at-off ratio in the synaptic array and (d) neuron circuits.

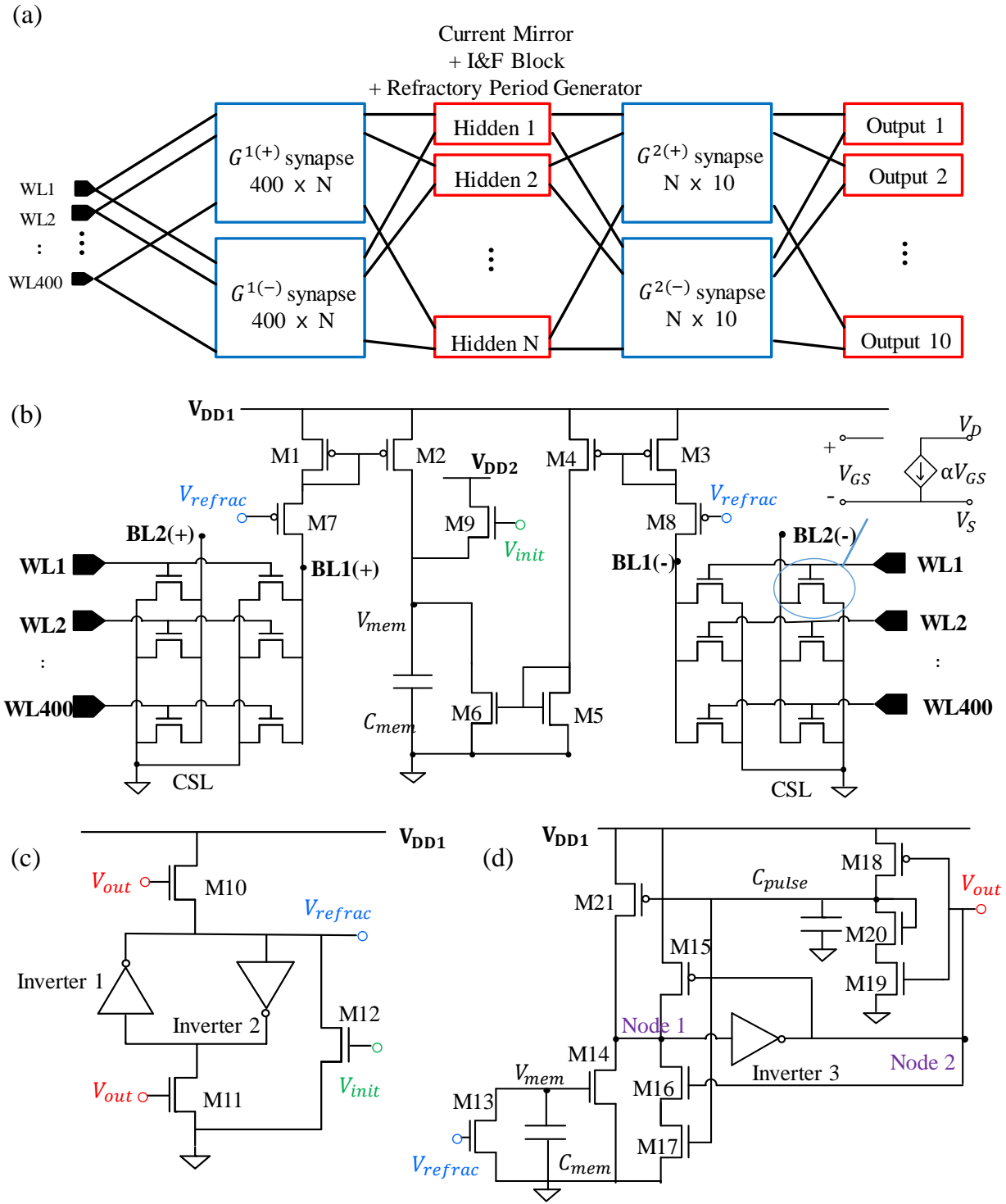

Fig. 5 (a) Conceptual diagram of the 2-layer network with N hidden neurons. Each weight value is represented by two synaptic devices, and each neuron circuit consists of a current mirror, an integrate-fire block, and a refractory period generator. Circuit diagram of the (b) synaptic array, current mirror, (c) refractory period generator and (d) integrate and fire block.

| Description | Components | Value |
|---|---|---|
| Width / Length of Transistor | M1 ~ M9, M21 | 0.5 μm / 2 μm |
| | M10 ~ M19, Inverter 1~3 | 0.5 μm / 0.5 μm |
| | M20 | 0.5 μm / 7 μm |
| Capacitance | $C_{mem, hidden}$ | 122 fF |
| | $C_{mem, output}$ | 77 fF |
| | $C_{pulse}$ | 2 pF |
| Supply Voltage | $V_{dd,1}$ | 2.5 V |
| | $V_{dd,2}$ | 0.4 V |
| Threshold Voltage of Neuron | $V_{th,hidden}$, $V_{th,out}$ | 0.92 V |

Table 2. Parameters of components used in the circuit-level simulation

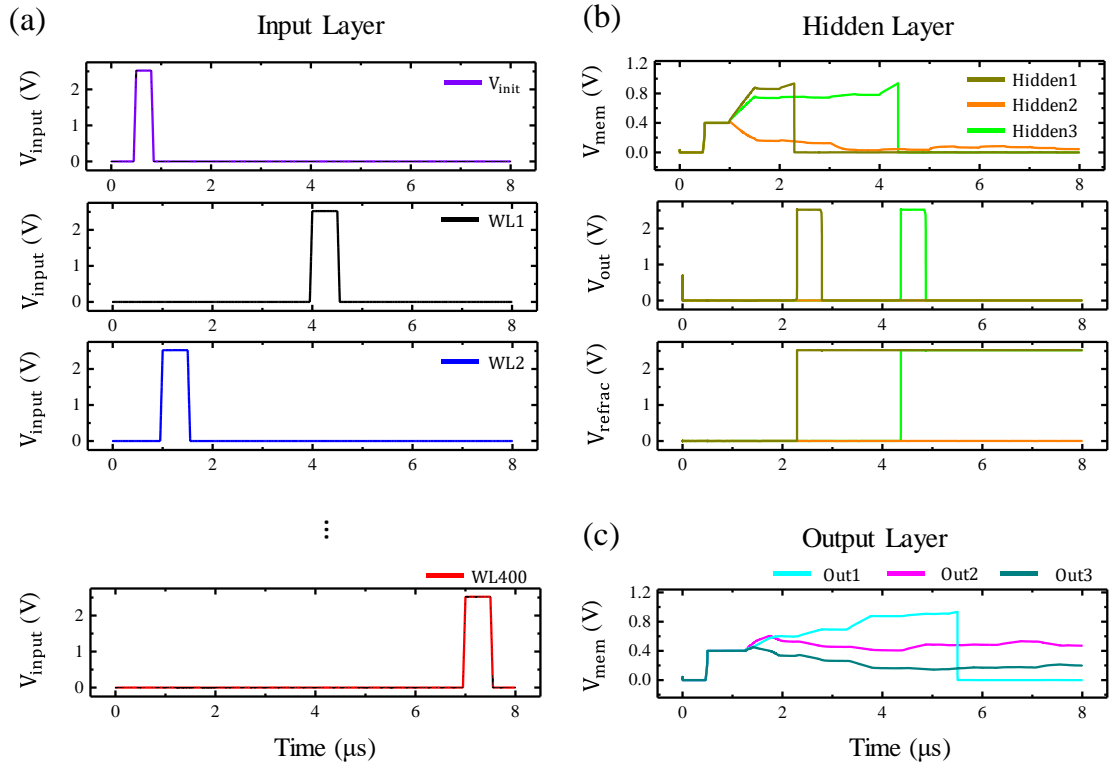

Figure 6. (a) Pulses fed into the input neuron shown in the time domain. (b) (Top) Evolution of the membrane voltage of hidden neurons. (Middle) Generated output pulse and (Bottom) refractory period by the neurons in hidden layer. (c) Evolution of the membrane voltage of output neurons. The answer predicted by the network is the class of output neuron 1. All results are simulated at circuit-level.

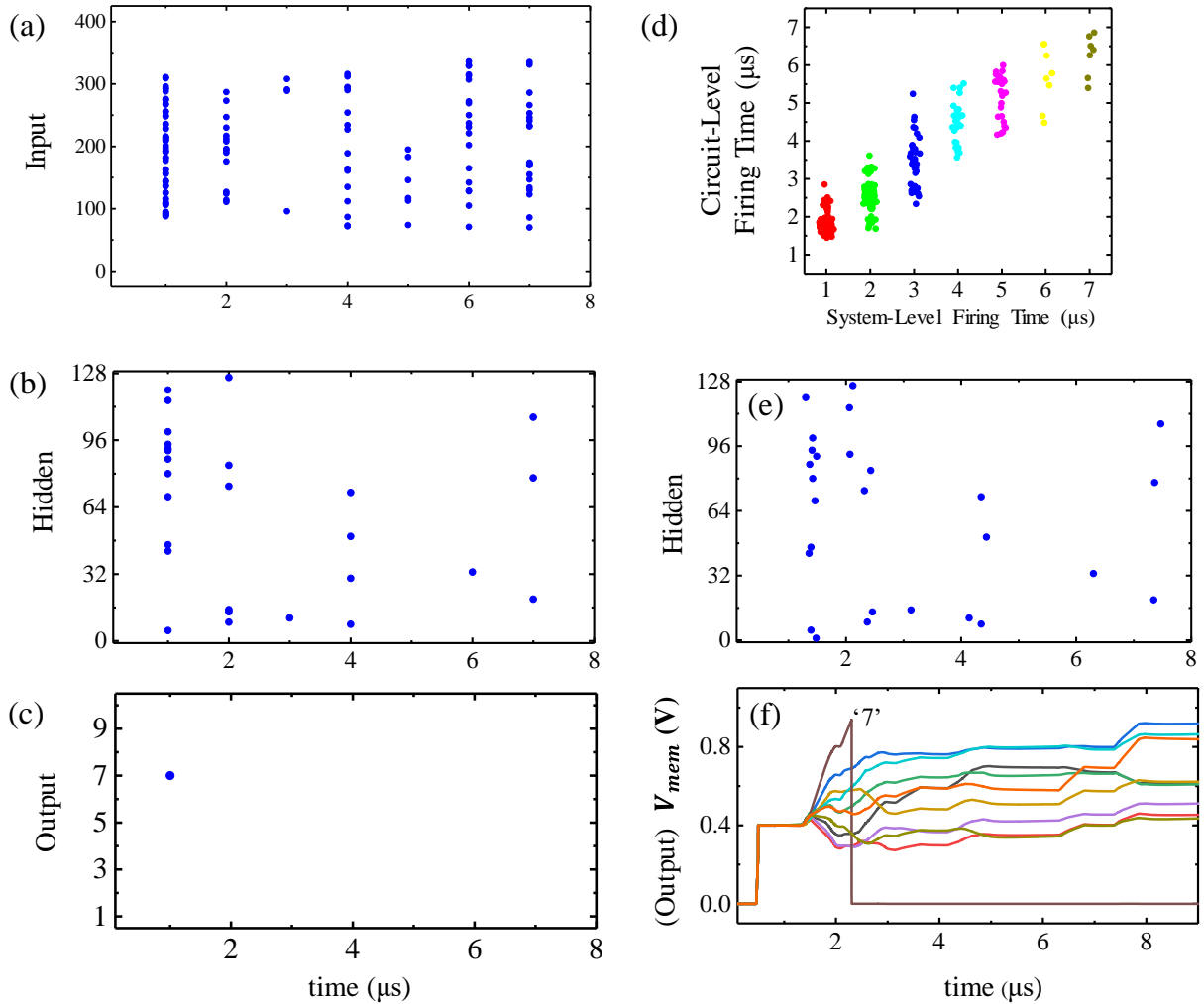

Figure 7. Raster plots of the spike timing of the (a) input, (b) hidden, and (c) output neuron in the 2-layer (400-128-10) SNN for randomly selected test data '7'. The x-axis represents the time in the system simulation, and the y-axis represents the index of each neuron. (d) Comparison of the firing time of winner neuron in the system-level (x-axis) and circuit-level (y-axis) simulation. (e) Raster plots of the hidden neuron when simulated in the circuit-level for the same network size and data as (b). (f) Evolution of the membrane voltage of output neurons in the circuit-level simulation.

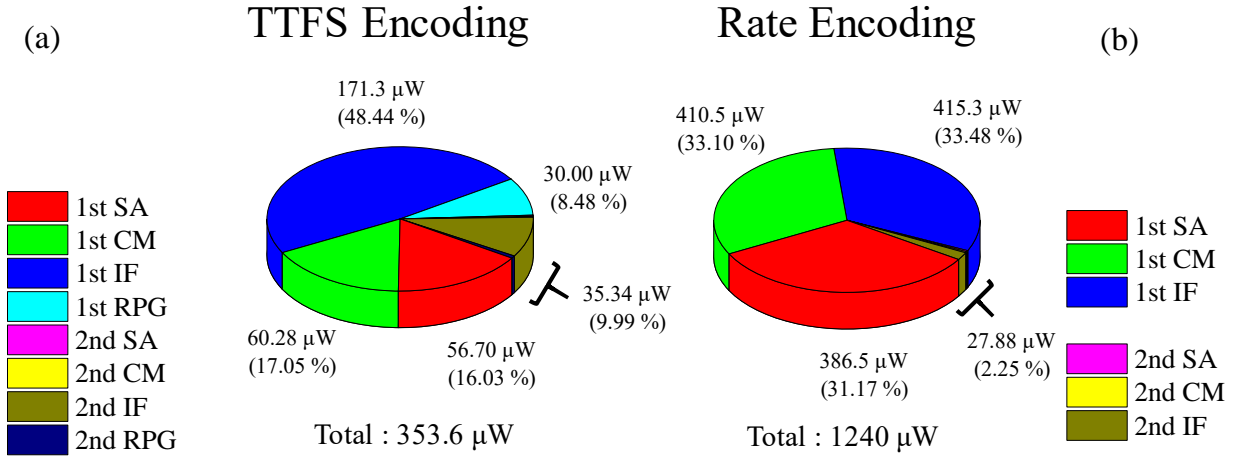

Figure 8. Pie chart of power consumption in (a) TTFS and (b) rate-encoded networks. The power consumption is analyzed by categorizing it into a synapse array (SA), current mirror (CM), integrate-and-fire (IF) and refractory generator (RPG) constituting the SNN.

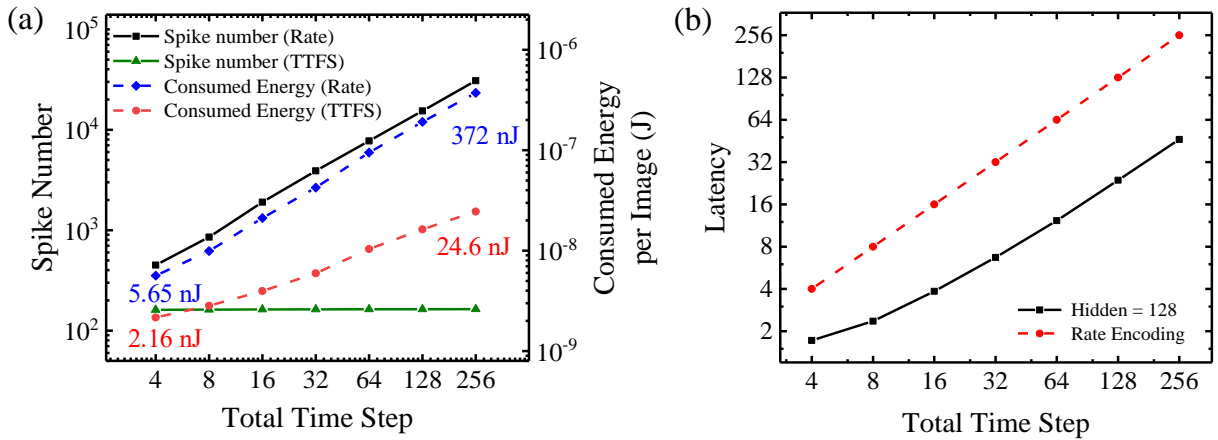

Figure 9. (a) Comparison between TTFS and rate-encoded network in terms of spike number (system-level) and consumed energy per image (circuit-level). Consumed energy was measured at various time steps on 10 randomly selected MNIST test sets. The simulation was conducted at various time steps. (b) The latency of TTFS network to make a decision. The dotted line indicates the total time step which is the time required for the rate encoded network to predict the answer.